\documentclass[letterpaper, 10 pt, conference]{ieeeconf}  

\IEEEoverridecommandlockouts

\usepackage{graphicx} % Required for inserting 
\usepackage{xcolor}

\makeatletter
\let\NAT@parse\undefined
\makeatother

\usepackage{hyperref}

\usepackage{amsmath}
\usepackage{outlines}
\usepackage{tensor}
\usepackage[hang,flushmargin,symbol]{footmisc}
\usepackage[capitalize]{cleveref}
\usepackage{flushend}
\crefname{section}{Sec.}{Secs.}
\Crefname{section}{Section}{Sections}

\crefname{figure}{Fig.}{Figs.}
\Crefname{figure}{Figure}{Figures}

\crefname{table}{Tab.}{Tabs.}
\Crefname{table}{Table}{Tables}

\crefname{algorithm}{Algo.}{Algos.}
\Crefname{algorithm}{Algorithm}{Algorithms}

\usepackage{xurl}

\usepackage[font=footnotesize]{caption}
\usepackage[font=footnotesize]{subcaption}
\usepackage{booktabs}
\usepackage{multirow}
\usepackage{multicol}
\usepackage{siunitx}
\usepackage{amssymb}

\usepackage{cite}

\usepackage{pifont}

\usepackage{amsmath}
\usepackage{amssymb}
\usepackage{placeins} % floatbarrier
\usepackage{pageslts} % page number
\usepackage{microtype}

\newcommand{\ourName}{CenterArt}
\newcommand{\ourSubName}{Joint Shape Reconstruction and 6-DoF Grasp Estimation of Articulated Objects}

\hypersetup{
    colorlinks=true,
    linkcolor=blue,
    filecolor=magenta,    
    citecolor=green,
    urlcolor=blue,
    pdftitle={\ourName: \ourSubName},
}

\title{\LARGE \bf
\ourName{}: \ourSubName
}

\author{Sassan Mokhtar, Eugenio Chisari, Nick Heppert, Abhinav Valada% <-this % stops a space
\thanks{Department of Computer Science, University of Freiburg, Germany}%
\thanks{This work was funded by Carl Zeiss Foundation with the ReScaLe
project.}
}

\begin{document}

\maketitle
\thispagestyle{empty}
\pagestyle{empty}

%%%%%%%%%%%%%%%%%%%%%%%%%%%%%%%%%%%%%%%%%%%%%%%%%%%%%%%%%%%%%%%%%%%%%%%%%%%%%%%%
\begin{abstract}
Precisely grasping and reconstructing articulated objects is key to enabling general robotic manipulation. In this paper, we propose \ourName{}, a novel approach for simultaneous 3D shape reconstruction and \mbox{6-DoF} grasp estimation of articulated objects. \ourName{} takes RGB-D images of the scene as input and first predicts the shape and joint codes through an encoder. The decoder then leverages these codes to reconstruct 3D shapes and estimate \mbox{6-DoF} grasp poses of the objects. We further develop a mechanism for generating a dataset of \mbox{6-DoF} grasp ground truth poses for articulated objects.
\ourName{} is trained on realistic scenes containing multiple articulated objects with randomized designs, textures, lighting conditions, and realistic depths. We perform extensive experiments demonstrating that \ourName{} outperforms existing methods in accuracy and robustness.
\end{abstract}

%%%%%%%%%%%%%%%%%%%%%%%%%%%%%%%%%%%%%%%%%%%%%%%%%%%%%%%%%%%%%%%%%%%%%%%%%%%%%%%%
\vspace{-3pt}
\section{Introduction}
\vspace{-1pt}
Manipulating articulated objects is crucial for many robotic applications such as household robots~\cite{schmalstieg2023learning,honerkamp2023n,honerkamp2024language}. However, before a robot can manipulate an object, it needs to acquire a grasp on a moveable part. Prior research addresses the \mbox{6-DoF} grasp pose generation (and articulation parameter estimation ~\cite{rofer2022kineverse, buchanan2023online}) problem for articulated objects through policy learning approaches utilizing reinforcement learning (RL)~\cite{mo2021where2act, zhang2023flowbot++, xu2022universal}. These approaches involve training a reinforcement learning agent to predict valid \mbox{6-DoF} grasp poses, which are then used to generate trajectories for object manipulation. However, RL-based methods demand significant amounts of data and training time. Furthermore, while they perform well under controlled conditions in simulations, they lack generalization to applications characterized by diverse scenes, illumination conditions, and noisy sensor observations.

Inspired by recent advances, we adopt a vision-based approach and propose \ourName{} for simultaneous 3D shape reconstruction and \mbox{6-DoF} grasp poses estimation of articulated objects. \ourName{} is an extension of \mbox{CenterGrasp}~\cite{chisari2023centergrasp}, a single-shot holistic grasp prediction approach for rigid-body objects.
To train \ourName{}, we set up two data generation procedures. First, we generate valid \mbox{6-DoF} grasp poses for articulated objects in an object-centric manner. Second, we use the Sapien simulator \cite{xiang2020sapien} to design and render realistic kitchen scenes including multiple articulated objects, leveraging previously generated grasps.

Our primary contributions can be summarized as follows:
\begin{itemize}%[topsep=0pt]
    \item The first approach for simultaneous 3D shape reconstruction and \mbox{6-DoF} grasp poses estimation of articulated objects.
    \item A dataset containing valid \mbox{6-DoF} ground truth grasp poses of articulated objects.
    \item Photo-realistic kitchen scenes consisting of several articulated objects.
\end{itemize}
%%%%%%%%%%%%%%%%%%%%%%%%%%%%%%%%%%%%%%%%%%%%%%%%%%%%%%%%%%%%%%%%%%%%%%%%%%%%%%%%
\begin{figure*}
    \centering
    \includegraphics[width=0.85\textwidth]{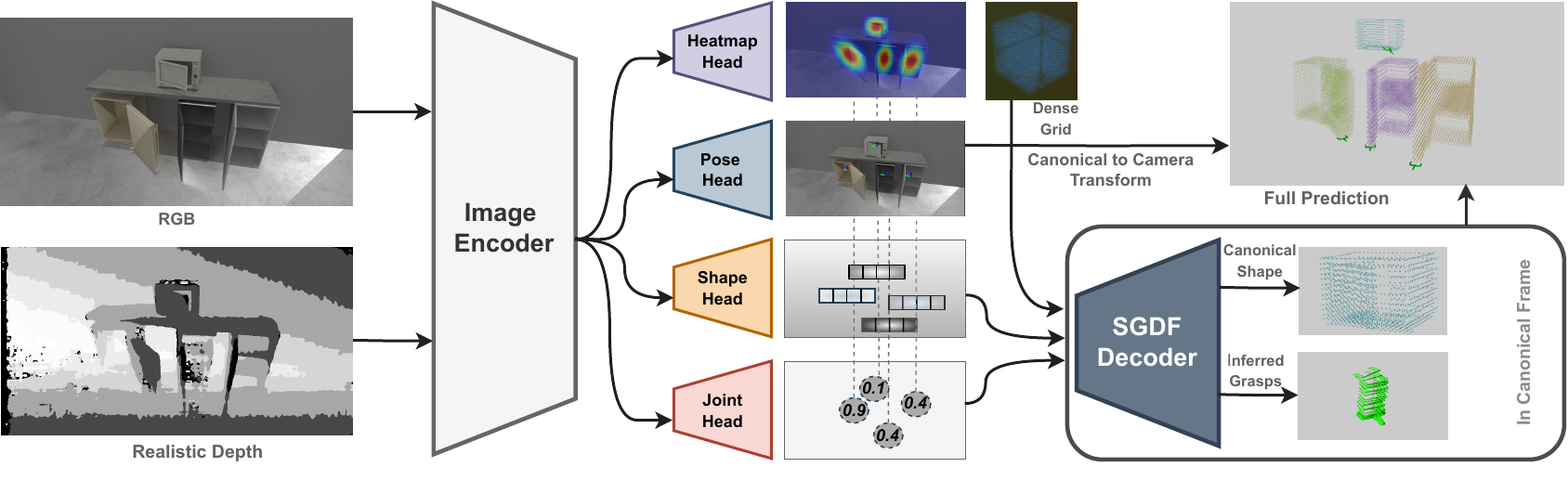}
    \caption{Overview of \ourName. First, input RGB-D images are encoded to predict object heatmaps, poses, shape codes, and joint codes in a per-pixel manner. Next, the peaks of heatmaps are used to detect the objects. The SGDF decoder then utilizes the predicted shape code and joint code to output the shape and grasp of detected objects. Finally, the estimated poses are used to transform the predicted 3D shapes and \mbox{6-DoF} grasps from the canonical frame to the camera frame.}
    \label{fig:overview}
\end{figure*}
%%%%%%%%%%%%%%%%%%%%%%%%%%%%%%%%%%%%%%%%%%%%%%%%%%%%%%%%%%%%%%%%%%%%%%%%%%%%%%%%
\vspace{-4pt}
\section{Related Work}
\vspace{-1pt}
Areas related to this work include center-based object detection, neural implicit representations for articulated objects, and grasp distance functions.

{\noindent \textit{Center-based Object Detection}:}
Inspired by single-stage object detectors such as YOLO~\cite{redmon2016you}, Zhou~\textit{et~al.} introduce CenterNet~\cite{zhou2019objects} which represents objects by a single point at their bounding box center, transforming object detection into a key point estimation problem. This approach improves accuracy and enhances predictions of object properties such as 6D pose estimation. \mbox{CenterSnap}~\cite{irshad2022centersnap} employs a center-based object detection method in a holistic manner to predict 6D poses and reconstruct 3D shapes. \ourName{} follows a similar approach to CenterSnap, using a point-based representation to detect and represent the complete 3D information (6D pose, 3D shape, and joint state) of articulated objects in the scene.

{\noindent \textit{Neural Implicit Representations for Articulated Objects}:}
Compared to rigid objects, articulated objects have more complex structures, making their tracking~\cite{heppert2022category} and reconstruction challenging. A-SDF (Articulated-SDF)~\cite{mu2021sdf} is one of the earliest works addressing this task with neural implicit representation. It represents articulated objects by disentangling codes for encoding shape and joint angle. CARTO~\cite{heppert2023carto} follows a holistic approach to detect, localize, and reconstruct articulated objects. Its decoder consists of two sub-decoders: a geometry decoder and a joint decoder. The shape reconstruction part of \ourName{} is closely related to CARTO, where an MLP is trained to concatenate the shape and joint code of the objects with sampled points to output the SDF value.

{\noindent \textit{Grasp Distance Functions}:}
Inspired by advances in neural implicit fields, Weng~\textit{et~al.}~\cite{weng2023neural} introduced Neural Grasp Distance Fields (NGDF), extending the concept of neural implicit distance functions to the domain of grasping tasks. NGDF predicts scalar distance metrics representing valid grasp poses for objects. The distance-based representation offered by NGDF can be interpreted as a cost function, which can be minimized through an optimization process. \mbox{CenterGrasp}~\cite{chisari2023centergrasp} proposes the Shape and Grasp Distance Function (SGDF), which is category-independent and handles multiple objects in the scene. We utilize SGDF from \mbox{CenterGrasp} and build upon it to develop a holistic approach for 3D shape reconstruction and \mbox{6-DoF} grasp pose estimation of articulated objects.

%%%%%%%%%%%%%%%%%%%%%%%%%%%%%%%%%%%%%%%%%%%%%%%%%%%%%%%%%%%%%%%%%%%%%%%%%%%%%%%%
\section{Technical Approach}
Given an RGB-D image of a scene with multiple articulated objects, the goal is to reconstruct the objects and predict valid grasp poses. \ourName{} consists of an image encoder that yields embedding vectors and joint states of each object in the scene, followed by a decoder that reconstructs the 3D shapes and determines valid \mbox{6-DoF} grasp poses (see \cref{fig:overview}).
\vspace{-4pt}
\subsection{Image Encoder}
\vspace{-3pt}
{\noindent \textit{Network Architecture}:} 
Inspired by the architecture of CenterSnap~\cite{irshad2022centersnap}, we first pass the RGB-D image separately through a ResNet50 \cite{he2016deep} to generate a low-resolution feature representation. Then, we concatenate the feature representations of RGB and depth images and feed them to a ResNet18-FPN backbone \cite{kirillov2019panoptic} to obtain a feature pyramid. Following a similar approach to SimNet \cite{kollar2022simnet}, we feed the resulting pyramid of features to specialized heads. We utilize the same structure as \mbox{CenterGrasp}~\cite{chisari2023centergrasp} for the heatmap, pose, and shape heads,  adding a joint head as the fourth head to the encoder. 
It predicts the joint state of the articulated object for each pixel of the downsampled map. To ensure a consistent representation of joint states, we determine the global maximum joint state and consider normalized joint states as ground truth labels.

{\noindent \textit{Losses}:} 
The total loss of the image encoder is given by
\vspace{-2pt}
\begin{equation*}
\begin{split}
\mathcal{L}_{\text{encoder}} &= w_{\text{heat}}\mathcal{L}_{\text{heat}} + w_{\text{pose}}\mathcal{L}_{\text{pose}} \\
&\quad+ w_{\text{shape}}\mathcal{L}_{\text{shape}} + w_{\text{joint}}\mathcal{L}_{\text{joint}}.
\end{split}
\end{equation*}
\vspace{-1pt}
Each loss is calculated using the mean squared error.

{\noindent \textit{Training}:}
The image encoder network is trained for 105 epochs using the ADAM optimizer with a learning rate of 1e$-$3. Additionally, color jitter augmentation is applied to the RGB images.
\vspace{-4pt}
\subsection{Shape and Grasp Decoder}
\vspace{-3pt}
The decoder aims to reconstruct 3D shapes and predict valid 6D grasp poses. Inspired by \mbox{CenterGrasp}~\cite{chisari2023centergrasp}, we utilize the shape and grasp distance function (SGDF) decoder to map shape code, joint state, and 3D coordinate to shape and grasp distances.

{\noindent \textit{Network Architecture}:}
Inspired by DeepSDF \cite{park2019deepsdf}, a deep feedforward multi-layer fully connected network is used for the decoder. The inputs of the network are shape code $\textbf{z}^s\in \mathbb{R}^{32}$, joint code $z^j\in \mathbb{R}^{1}$, and a 3D point $\textbf{x}\in \mathbb{R}^{3}$. Utilizing an 8-layer perceptron with 512 neurons at each layer, in the first layer shape code and 3D points are fed to the decoder. The joint code appends to the second layer. Moreover, the shape code, joint code, and the 3D point append to the output of the fourth layer. The activation function of hidden layers is ReLU, while the activation function of the last layer is the hyperbolic tangent.

{\noindent \textit{Losses}:}
For the SDF values, the clamp function is used, which constrains its input value. The loss is then defined by the $L1$ loss of the clamped SDF values. To have a uniform loss for the translation and rotation components of the grasp pose loss, we follow \cite{sundermeyer2021contact} to process the target and predicted grasp poses to represent the 6D grasp pose with five 3D points $\textbf{o}^{gp}, \hat{\textbf{o}}^{gp}\in \mathbb{R}^{3\times 5}$. Then, the grasp pose loss is simply the $L1$ distance between the target and predicted points. The third component of the loss is designed to regularize the shape codes. The total loss of the decoder is then given by
\vspace{-2pt}
\begin{equation*}
\mathcal{L}_{decoder} = w_{SDF}\mathcal{L}_{SDF} + w_{grasp}\mathcal{L}_{grasp} + w_{code}\mathcal{L}_{code}.
\end{equation*}
\vspace{-1pt}
{\noindent \textit{Training}:}
Each articulated object is paired with 7 to 10 different joint states. For every object, we sample one joint state and include the corresponding object joint state pair in a validation set. Then, for each object joint state pair, we sample 100,000 points with corresponding SDF values. The grasp distance label is computed for each point by finding the closest ground-truth grasp. The SGDF network is trained for 600 epochs, with ADAM optimizer and step-based decay learning rate scheduler between 1e$-$3 to 25e$-$5. Additionally, dropout with probability $0.2$ and weight normalization are applied for regularization.
\vspace{-4pt}
\subsection{Full \ourName{} Inference}
\vspace{-3pt}
Given an RGB-D input, the image encoder produces per-pixel predictions for object heatmap, 6D pose, shape code, and joint code. Each peak in the heatmap is assumed to be the object's center, which is treated as the representative of the object. We then extract the 6D pose, shape code, and joint code corresponding to each object center to input into the SGDF decoder for per-object prediction.
For each object, we concatenate the shape code and joint code to create a latent code specific to the \textit{object-joint state} pair. This latent code, along with sampled 3D coordinates in a dense grid, is fed into the decoder to predict an SDF value and a \mbox{6-DoF} grasp for each sampled point. We identify object surface points and a set of valid grasp poses by considering iso-surfaces $SGDF(.)=0$. Finally, the 6D pose of the object is utilized to transform the remaining points and grasps from their canonical frame to the camera frame.
%%%%%%%%%%%%%%%%%%%%%%%%%%%%%%%%%%%%%%%%%%%%%%%%%%%%%%%%%%%%%%%%%%%%%%%%%%%%%%%%
\vspace{-5pt}
\begin{figure}[h]
    \centering
    \includegraphics[width=0.48\textwidth]{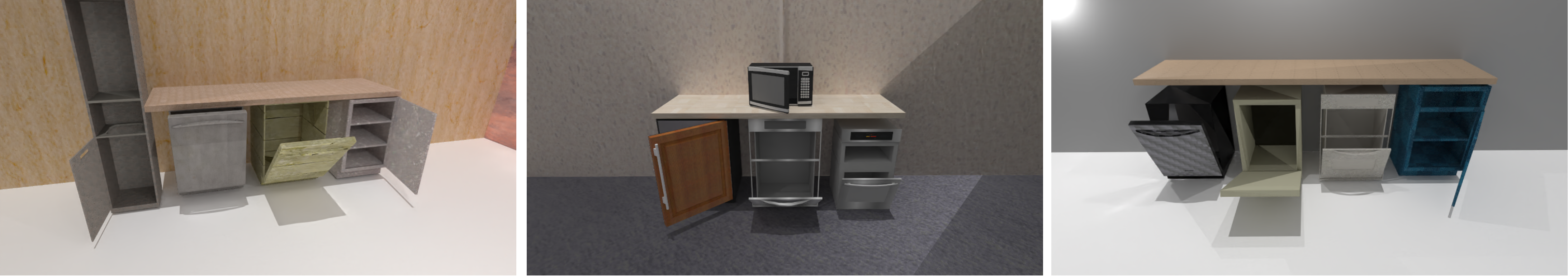}
    \caption{Generated kitchen scenes}
    \label{fig:scene}
\end{figure}

%%%%%%%%%%%%%%%%%%%%%%%%%%%%%%%%%%%%%%%%%%%%%%%%%%%%%%%%%%%%%%%%%%%%%%%%%%%%%%%%
\vspace{-20pt}
\subsection{Dataset Generation}
\vspace{-3pt}
{\noindent \textit{Object-Centric \mbox{6-DoF} Grasp Generation}:}
Initially, 82 articulated objects were collected from the PartNet-Mobility dataset \cite{mo2019partnet}, covering five different categories: Microwave, Oven, Refrigerator, Dishwasher, and Storage Furniture.
After performing a preprocessing step on the collected data, we consider ten different joint states for each object, which is obtained by evenly splitting the distance between the minimum and maximum joint state. Subsequently, for each \textit{object-joint state} pair, a point cloud of the articulated link is generated. Then, the positions of ground-truth grasps are a subset of the generated point cloud.

To reduce the full rotation manifold and thus, speed-up data generation, we utilize the articulation axis and the joint state. We calculate the set of all valid orientations corresponding to three edges and the possible handle of the articulated link of the object. Then, for each point in the point cloud, we sample one orientation among a set of all valid orientations and form a candidate grasp pose. We then evaluate the candidate grasp in PyBullet simulator \cite{coumans2016pybullet}. If the flying gripper can move the articulated link successfully, then the candidate grasp is regarded as a valid grasp pose. 

We generate and store between $100$ to $500$ grasp poses for each \textit{object-joint state} pair. Finally, all generated grasps were manually verified to exclude \textit{object-joint state} pairs with insufficient grasp labels or where the labels did not cover all areas of interest in the link. Overall, $375,266$ grasp labels were generated for $766$ \textit{object-joint state} pairs. 

%%%%%%%%%%%%%%%%%%%%%%%%%%%%%%%%%%%%%%%%%%%%%%%%%%%%%%%%%%%%%%%%%%%%%%%%%%%%%%%%
{\noindent \textit{Full Scene Generation}:}
To train the encoder and evaluate the full pipeline, we set up a generation process for realistic kitchen scenes with single or multiple articulated objects. For training, we create about 25,000 random scenes. Each scene is rendered from four random camera poses, resulting in approximately 100,000 RGB-D images and labels. Rendering is done using the Sapien raytracing-based renderer \cite{xiang2020sapien}, with realistic depth images \cite{zhang2023close}. Object heatmaps are generated by fitting a Gaussian to the ground truth masks, with the peaks indicating object locations in the image. (see \cref{fig:scene})
%%%%%%%%%%%%%%%%%%%%%%%%%%%%%%%%%%%%%%%%%%%%%%%%%%%%%%%%%%%%%%%%%%%%%%%%%%%%%%%%

\begin{table}
\centering
\begin{tabular}{llll|ll}
\hline
\multirow{2}{*}{\textbf{Scene}} & \multirow{2}{*}{\textbf{Method}} & \multicolumn{2}{l|}{\textbf{GT-depth}} & \multicolumn{2}{l}{\textbf{Noisy-depth}} \\ \cline{3-6} 
 &  & \textbf{SR} & \textbf{RSR} & {SR} & \textbf{RSR} \\ \hline
{Single Obj} & \multicolumn{1}{l|}{{UMPNet}} & 0.24 & 0.53 & 0.00 & 0.01 \\
{Single Obj} & \multicolumn{1}{l|}{{\ourName}} & \textbf{0.52} & \textbf{0.95} & \textbf{0.53} & \textbf{0.97} \\
{Single Obj} & \multicolumn{1}{l|}{{\ourName{} + ICP}} & 0.52 & 0.75 & 0.51 & 0.75 \\ \hline
{Multiple Objs} & \multicolumn{1}{l|}{{\ourName}} & 0.26 & \textbf{0.72} & \textbf{0.51} & \textbf{0.94} \\
{Multiple Objs} & \multicolumn{1}{l|}{{\ourName{} + ICP}} & \textbf{0.33} & 0.66 & 0.29 &  0.70\\ \hline
\end{tabular}

\caption[short caption]{Evaluation of \mbox{6-DoF} grasp pose estimation for unseen objects. \\ SR = Success Rate ($\uparrow$), RSR = Relaxed Success Rate ($\uparrow$)}
\label{tab:res}
\end{table}

%%%%%%%%%%%%%%%%%%%%%%%%%%%%%%%%%%%%%%%%%%%%%%%%%%%%%%%%%%%%%%%%%%%%%%%%%%%%%%%%
\vspace{-4pt}
\section{Experimental Results}
We conducted experiments to evaluate the performance of \ourName{} and compare against UMPNet~\cite{xu2022universal}, a state-of-the-art baseline for grasp estimation of articulated objects. UMPNet is an RL-based approach that estimates the \mbox{6-DoF} grasp of articulated objects and predicts manipulation trajectories. Since \ourName{} only estimates grasp poses, we used the corresponding part of UMPNet for comparison.
UMPNet uses ground truth depth for training. To ensure a fair comparison, noise-free depths were provided to UMPNet. Additionally, UMPNet is trained on simple scenes with a single object and a floor. Thus, we alter our data generation process to exclude walls and multiple objects. For evaluating \ourName{}, two scene variations are considered: scenes with a single object in a room, similar to UMPNet scenes, and more complex kitchen scenes with multiple objects.

We utilize two metrics for evaluation. The \mbox{Success Rate (SR)} is defined as the successful movement of the articulated joint for at least 10 degrees. If the difference between the maximum joint state and the current joint state is less than 45 degrees, the goal is to close the joint; otherwise, the goal is to open the joint. Another metric is the \mbox{Relaxed Success Rate (RSR)}, which determines if the predicted grasp pose is close enough to any ground truth grasps. In this metric, the prediction is regarded as successful if the minimum Euclidean distance between the predicted grasp position and any grasp label is less than $~10\%$ of the initial distance. The experiments are performed in the Sapien simulator \cite{xiang2020sapien} with a flying gripper.

We present the results in \cref{tab:res}. 
On the single object scenes, \ourName{} doubles the success rate over the baseline UMPNet ($52\%$ compared to $24\%$).
Additionally, \ourName{} consistently performs well on both ground truth and noisy depth images, whereas UMPNet fails with noisy depth images. Even in complex kitchen scenes containing multiple objects, \ourName{} still performs better than UMPNet in simple scenes with a single object. Finally, it is worth noting that while refining the predicted poses with ICP contributes to better results in \mbox{CenterGrasp}~\cite{chisari2023centergrasp}, as shown in \cref{tab:res}, it does not result in consistent improvement of \ourName{}.

\section{Conclusion}
In this work, we introduced \ourName{}, a vision-based approach that simultaneously performs shape reconstruction and \mbox{6-DoF} grasp estimation of articulated objects. Additionally, we generated a dataset of valid \mbox{6-DoF} grasp poses and realistic kitchen scenes with multiple articulated objects. Our experiments demonstrate that \ourName{} improves the success rate of state-of-the-art baseline by $28\%$. Moreover, \ourName{} achieves a consistent performance across various scenarios, including those with noisy depth and realistic kitchen scenes, highlighting its robustness in practical settings.

\newpage
\bibliographystyle{ieeetr}
\bibliography{root}

\end{document}